\documentclass[draftcls]{IEEEtran}
\usepackage[caption=false,font=normalsize,labelfont=sf,textfont=sf]{subfig}
\usepackage{url}
\usepackage{verbatim}
\usepackage{cite}
\ifCLASSINFOpdf
\else
\fi
\usepackage{microtype}
\usepackage{array}
\usepackage{mdwmath}
\usepackage{mdwtab}
\usepackage{graphicx}
\usepackage{euscript}
\usepackage{amsfonts}
\usepackage{bigstrut}
\usepackage{tabularx}
\usepackage{cite}
\usepackage{amsmath}
\usepackage{algorithm}
\usepackage[noend]{algpseudocode}
\makeatletter
\def\BState{\State\hskip-\ALG@thistlm}
\makeatother
\usepackage{enumitem}
\usepackage{upgreek}
\usepackage{dsfont}
\usepackage{amsthm}
\usepackage{amssymb}
\usepackage{textcomp}
\usepackage{gensymb}
\usepackage{xcolor}
\usepackage{mathtools}

\usepackage{hyperref}
\usepackage{epstopdf} 
\usepackage{bbm}
\DeclareMathOperator*{\argmin}{argmin} 
\usepackage{microtype}
\usepackage{booktabs}

\usepackage{multirow}
\usepackage{threeparttable}  

\title{
A Multi-Drone Multi-View Dataset and Deep Learning Framework for Pedestrian Detection and Tracking
}

\author{Kosta Dakic, Kanchana Thilakarathna, Rodrigo N. Calheiros, and Teng Joon Lim\\
}

\begin{document}
	\onecolumn

\maketitle
\begin{abstract}
Multi-drone surveillance systems offer enhanced coverage and robustness for pedestrian tracking, yet existing approaches struggle with dynamic camera positions and complex occlusions. This paper introduces MATRIX (Multi-Aerial TRacking In compleX environments), a comprehensive dataset featuring synchronized footage from eight drones with continuously changing positions, and a novel deep learning framework for multi-view detection and tracking. Unlike existing datasets that rely on static cameras or limited drone coverage, MATRIX provides a challenging scenario with 40 pedestrians and a significant architectural obstruction in an urban environment. Our framework addresses the unique challenges of dynamic drone-based surveillance through real-time camera calibration, feature-based image registration, and multi-view feature fusion in bird's-eye-view (BEV) representation. Experimental results demonstrate that while static camera methods maintain over 90\% detection and tracking precision and accuracy metrics in a simplified MATRIX environment without an obstruction, 10 pedestrians and a much smaller observational area, their performance significantly degrades in the complex environment. Our proposed approach maintains robust performance with $\sim$90\% detection and tracking accuracy, as well as successfully tracks $\sim$80\% of trajectories under challenging conditions. Transfer learning experiments reveal strong generalization capabilities, with the pretrained model achieving much higher detection and tracking accuracy performance compared to training the model from scratch. Additionally, systematic camera dropout experiments reveal graceful performance degradation, demonstrating practical robustness for real-world deployments where camera failures may occur. The MATRIX dataset and framework provide essential benchmarks for advancing dynamic multi-view surveillance systems.
\end{abstract}


\section{Introduction}~\label{Section_I}

\IEEEPARstart{M}{ulti-camera} systems have dramatically improved our ability to track and analyze multi-target movements in complex environments~\cite{AMOSA2023126558}. Multi-camera setups overcome key limitations of single-camera systems by providing wider coverage, handling occlusions, and enabling more robust tracking through the fusion of multiple viewpoints~\cite{Olagoke2020}. Nevertheless, the increasing deployment of drone-based surveillance systems introduces new challenges that remain largely unaddressed in current literature. Traditional multi-view approaches assume fixed camera positions, which simplify feature correlation and trajectory prediction. In contrast, drone-based systems must contend with continuously changing camera positions, dynamic occlusions, and the need for real-time view registration~\cite{Akbari2021}. Multi-drone systems have emerged as an effective solution to overcome the limitations of single-view systems~\cite{10008047}, particularly in scenarios involving occlusions and rapid target movements. These systems offer enhanced coverage and viewpoint diversity, enabling more robust tracking and detection performance~\cite{zhu2021detection, Campus}. However, coordinating multiple drone perspectives introduces unique technical challenges, including view synchronization and dynamic camera calibration.

Recent advances in deep learning have revolutionized computer vision tasks, from autonomous driving~\cite{grigorescu2020survey} to medical imaging~\cite{litjens2017survey} and surveillance systems~\cite{liu2020deep}, with a particularly significant impact on multi-view detection and tracking~\cite{chen2017multi}. These developments have enabled new fusion approaches, from early fusion methods that combine raw sensor data~\cite{teepe2023earlybird} to late fusion techniques that integrate features at region proposal or detection levels~\cite{Fadadu_2022_WACV, 9879824}. While traditional methods relied on probabilistic occupancy maps and geometric constraints~\cite{4359319, Berclaz2011}, modern approaches leverage deep neural networks for feature extraction and matching~\cite{hou2020multiview, 101145}. Despite these advances, existing solutions struggle with the dynamic nature of drone-based surveillance, where camera positions and orientations change continuously. Recent architectures have made significant progress in addressing these challenges. Advanced feature fusion approaches have improved the field of multi-view perception~\cite{101145, BEVformer}, with modern models excelling at capturing comprehensive scene understanding and demonstrating superior performance in aggregating information across multiple views. Furthermore, these architectures focus on efficiency and parallelization, maintaining computational performance suitable for real-time applications, a significant challenge that must be overcome to allow for efficient mobile edge computing.

Recent research has shown that birds-eye-view (BEV) representations can significantly improve multi-view perception tasks~\cite{hou2020multiview}. BEV projections provide a unified coordinate system for feature fusion and object localization~\cite{BEVformer}. However, existing BEV approaches primarily focus on scenarios with fixed camera configurations, and adapting these techniques to dynamic drone networks requires novel solutions for continuous view registration and feature alignment~\cite{brown1992survey, 7769090}. Data association remains a fundamental challenge in drone-based surveillance networks~\cite{10008047}. Drone-based systems introduce unique technical difficulties: cameras mounted on drones create continuously changing viewpoints, which complicates feature correlation and trajectory prediction compared to fixed camera setups~\cite{8296962}. Additional challenges include handling varying object scales due to changing drone altitudes and distances~\cite{Mueller2017, du2018unmanned}, as well as meeting the strict computational and real-time processing requirements of aerial platforms~\cite{8578694}.

To advance drone-based surveillance capabilities, we propose a comprehensive multi-view detection and tracking pipeline specifically designed for dynamic drone-based scenarios. Our approach introduces several key innovations: real-time perspective transformation to BEV representation, feature-based image registration for robust view alignment, and multi-view feature fusion. Unlike existing approaches~\cite{hou2020multiview, 101145, teepe2023earlybird, teepe2024lifting}, which primarily focus on static camera setups, our pipeline actively handles the challenges of moving cameras through continuous view registration and dynamic feature correlation. Our experimental results demonstrate that the proposed pipeline performs better in dynamic camera scenarios than existing methods, particularly in scenarios with a dense number of tracked objects and in maintaining consistent tracks through occlusions and viewpoint changes. 

 We additionally demonstrate our MATRIX dataset, which is generated using Unreal Engine 5~\cite{unrealengine5} with AirSim~\cite{airsim} for precise drone control, provides a challenging new benchmark that addresses critical gaps in existing multi-view datasets. Unlike static camera datasets such as WILDTRACK~\cite{8578626} and MultiviewX~\cite{hou2020multiview}, which feature fixed camera positions, or existing drone datasets like MDMT~\cite{10008047} and MUMO~\cite{9913874} that are limited to 2-4 cameras, MATRIX combines the benefits of comprehensive multi-view coverage (8 drones) with dynamic camera movements. The dataset's complexity is further enhanced through two distinct scenarios: a simple environment with minimal occlusions and a complex scenario featuring dense crowds (40+ pedestrians) with significant architectural obstructions. This combination of dynamic viewpoints, high pedestrian density, and substantial occlusions creates substantially more challenging conditions than existing benchmarks, where state-of-the-art methods that perform well on static multi-view datasets show significant performance degradation. The MATRIX dataset thus encourages the development of more robust algorithms specifically designed for dynamic urban surveillance scenarios.

We believe this work represents a significant step forward in bridging the gap between traditional static multi-view systems and the dynamic requirements of modern drone-based surveillance. The main contributions of this paper are as follows:
\begin{itemize}
\item Introduction of the MATRIX Dataset, featuring synchronized footage from eight drones in an urban environment with comprehensive annotations for detection and tracking, available at~\url{https://github.com/KostaDakic/MATRIX/tree/main}.
\item Development of a novel dynamic camera calibration system that provides continuous updates of extrinsic parameters as drones move through the surveillance area.
\item Design of an efficient multi-view feature fusion pipeline that maintains high detection and tracking performance while adapting to rapid scene changes and viewpoint variations.
\item Comprehensive evaluation demonstrating robust performance in both simple and complex environments with major occlusions, strong generalization capabilities through transfer learning to modified scenarios, and graceful degradation under systematic camera dropout
\end{itemize}

The remainder of this article is organized as follows. Sec.\ref{Section_II} introduces the background and related work in drone-based datasets, multi-view detection, and tracking. Sec.\ref{Section_III} presents our MATRIX dataset and its key characteristics, along with our annotation methodology. Sec.\ref{Section_IV} details our proposed methodology, including the dynamic camera calibration system and our multi-view feature fusion pipeline. Sec.\ref{Section_V} presents comprehensive experimental results and analysis, comparing our approach with state-of-the-art methods. Finally, Sec.~\ref{Section_VI} concludes the paper and discusses future research directions in dynamic multi-view surveillance.

\section{Background and Related work}~\label{Section_II}
This section summarizes relevant research in object detection and tracking, with a focus on multi-view approaches and their relevance to the MATRIX dataset.

\subsection{Evolution of Object Detection and Tracking}
Object detection and tracking have evolved significantly, from traditional methods using hand-crafted features to modern deep-learning approaches. Bewley et al.~\cite{8296962} enhanced the SORT algorithm by incorporating appearance information, improving tracking through occlusions. The introduction of Transformers to computer vision marked a paradigm shift, with DETR~\cite{carion2020end} simplifying the detection pipeline by eliminating hand-designed components. Zhu et al.~\cite{zhu2021deformable} further improved upon this with Deformable DETR, addressing issues of slow convergence and small object detection.

Multi-object tracking remains a fundamental challenge in computer vision. Several notable approaches have emerged to tackle different aspects of this challenge. TrackFormer, proposed by Meinhardt et al.\cite{9879668}, introduced an end-to-end Transformer-based model for simultaneous detection and tracking, demonstrating how modern architectures can unify traditionally separate tracking components. For edge computing applications, MobileDR by Anand et al.\cite{9557791} optimized tracking for low-power devices using depthwise separable convolutions and multi-sensor fusion, making real-time tracking feasible on resource-constrained platforms. In the realm of sensor fusion, Wang et al.~\cite{WANG2022} developed DeepFusionMOT, a real-time 3D MOT framework that effectively combines camera and LiDAR data through a deep association mechanism. Their approach leverages cameras' ability to detect distant objects and LiDAR's precise depth measurements, achieving state-of-the-art performance on standard tracking benchmarks while maintaining real-time processing capabilities.

Image registration has been crucial in bridging single-view and multi-view perception systems. Traditional feature-based methods like SIFT~\cite{lowe2004distinctive} and SURF~\cite{bay2006surf} laid the groundwork for robust keypoint detection and matching. More recently, deep learning approaches have significantly advanced the field. SuperGlue~\cite{sarlin2020superglue} demonstrated how graph neural networks could improve feature matching by leveraging learned contextual information. In the context of aerial imagery, GFTT-AffNet-HardNet~\cite{mishkin2018repeatability} has shown particular promise by combining affine-covariant feature detection with robust descriptors, making it well-suited for the perspective changes common in drone footage. These advancements in image registration have been instrumental in enabling accurate feature alignment across multiple views, a critical requirement for robust multi-view tracking systems.

\subsection{Multi-Camera and Multi-View Approaches}
The evolution of multi-camera tracking systems has progressed through several key developments, with a particular focus on maintaining object identity consistency across different views. A significant breakthrough came from Ristani and Tomasi~\cite{8578730}, who demonstrated the effectiveness of Convolutional Neural Networks (CNNs) in both Multi-Target Multi-Camera Tracking (MTMCT) and Person Re-Identification (Re-ID), establishing robust methods for feature learning across multiple camera perspectives.

Building on these foundations, multi-view detection and tracking emerged as a more sophisticated approach, enabling comprehensive scene understanding through the integration of multiple viewpoints. The seminal work of Fleuret et al.~\cite{4359319} established a framework combining probabilistic occupancy maps (POM) with global trajectory optimization, laying the groundwork for modern multi-view systems. This foundation was further advanced by Hou et al.~\cite{hou2020multiview} with their MVDet architecture, which specifically addressed the challenges of occlusion and crowded scenes through innovative feature map perspective transformation and bird's-eye view (BEV) representation. Their subsequent work, MVDeTr~\cite{101145}, enhanced this approach by incorporating a shadow transformer to better handle projection distortions. Complementing these advances, Xu et al.~\cite{7780830} expanded the tracking methodology by integrating multiple cues beyond ground plane locations, including appearance features, motion coherence, and geometric proximity, creating a more robust framework for multi-view tracking association. While these multi-view approaches have focused primarily on accuracy improvements, our prior work~\cite{dakic2024} addressed the orthogonal challenge of communication efficiency in distributed multiview systems through semantic-guided masking and masked autoencoders. However, these communication-efficient approaches assumed static camera configurations, leaving the challenges of dynamic drone-based surveillance unaddressed.

\subsection{Recent Advancements in Multi-View Tracking}
Building on these foundations, Teepe et al. proposed EarlyBird~\cite{teepe2023earlybird}, a framework performing early fusion of camera views in BEV space for joint detection and tracking. Their follow-up work~\cite{teepe2024lifting} compared different lifting algorithms for projecting multi-view features into BEV representation, introducing temporal aggregation for improved tracking. In the domain of collaborative perception, Zhou et al.~\cite{zhou2024centercoop} introduced CenterCoop, leveraging BEV space for efficient feature aggregation across vehicle and infrastructure sensors. To address distortion issues in multi-view aggregation, Wang et al.~\cite{10031058} proposed MVTT, a novel feature transformation approach.

\subsubsection{Drones}
Recent advances in drone-based surveillance systems have sparked significant research interest in addressing the unique challenges of aerial object detection and tracking. These challenges primarily stem from the dynamic nature of aerial platforms, varying perspectives, and the need to handle small objects at different scales. The literature in this domain can be categorized into three main areas: single-drone detection systems, multi-drone tracking frameworks, and dataset contributions.

Early approaches to drone-based detection primarily focused on 2D object detection, limiting their ability to provide accurate spatial understanding. Addressing this limitation, Wang et al.~\cite{10044977} introduced DVDET. This dual-view detection system performs detection in both 2D image and 3D physical space. In contrast, Wen et al.~\cite{DroneCrowd} proposed STNNet, which integrates density map estimation with localization to handle the challenges of crowded scenes. While these works made significant advances in single-drone detection, addressing challenges such as severe scale variations and geometric distortions through innovative approaches like DVDET's bin categorization for altitude estimation and STNNet's space-time neighbor-aware architecture, they remain limited by their single-viewpoint perspective.

The evolution toward multi-drone systems has introduced new challenges in identity association and occlusion handling. Liu et al.~\cite{10008047} address these challenges through their Multi-matching Identity Authentication network (MIA-Net), which represents a significant advancement in multi-drone tracking. Unlike previous approaches that treated each drone's view independently, MIA-Net introduces a local-global matching algorithm that effectively leverages the complementary information from multiple views. Transformer-based architectures have recently shown promise in multi-drone tracking applications. Chen et al.~\cite{10144283} propose TransMDOT, which leverages a Vision Transformer backbone to learn rich associations between object templates across multiple drone views. Their cross-drone mapping module represents a notable advancement in handling target re-identification when objects leave one drone's field of view. While effective, their approach primarily focuses on single object tracking, leaving room for improvement in multi-object scenarios typical in urban surveillance. Earlier work by Zhang et al.~\cite{9298794} introduced the Agent Sharing Network (ASNet) alongside the MDOT dataset, establishing a foundation for multi-drone tracking research. Their self-supervised template sharing mechanism and view-aware fusion scheme demonstrated the potential of collaborative perception across multiple drones. However, their approach assumes relatively stable drone positions, limiting its effectiveness in scenarios with rapidly changing viewpoints.

Dataset contributions have played a crucial role in advancing the field. Wang et al.'s~\cite{10044977} AM3D-Sim and AM3D-Real datasets address the critical need for comprehensive 3D detection data from aerial perspectives. Their demonstration that models pre-trained on simulation data can enhance real-world performance opens new possibilities for training robust detection systems. However, as shown in Table~\ref{Drone_DS_Table}, existing drone datasets are predominantly focused on single-drone scenarios or limited multi-drone setups with just 2-4 cameras, such as MDMT and MUMO. Meanwhile, Table~\ref{MV_DS_Table} illustrates that established multi-view datasets like WILDTRACK and MultiviewX, while offering comprehensive coverage with 6-7 cameras, are limited to static camera setups. This creates a significant gap in the evaluation of algorithms under dynamic conditions, as none of the existing datasets combine the benefits of both multi-view coverage and drone mobility.

\begin{figure*}[ht]
    \normalsize
	\centering
   	\includegraphics[width=\textwidth]{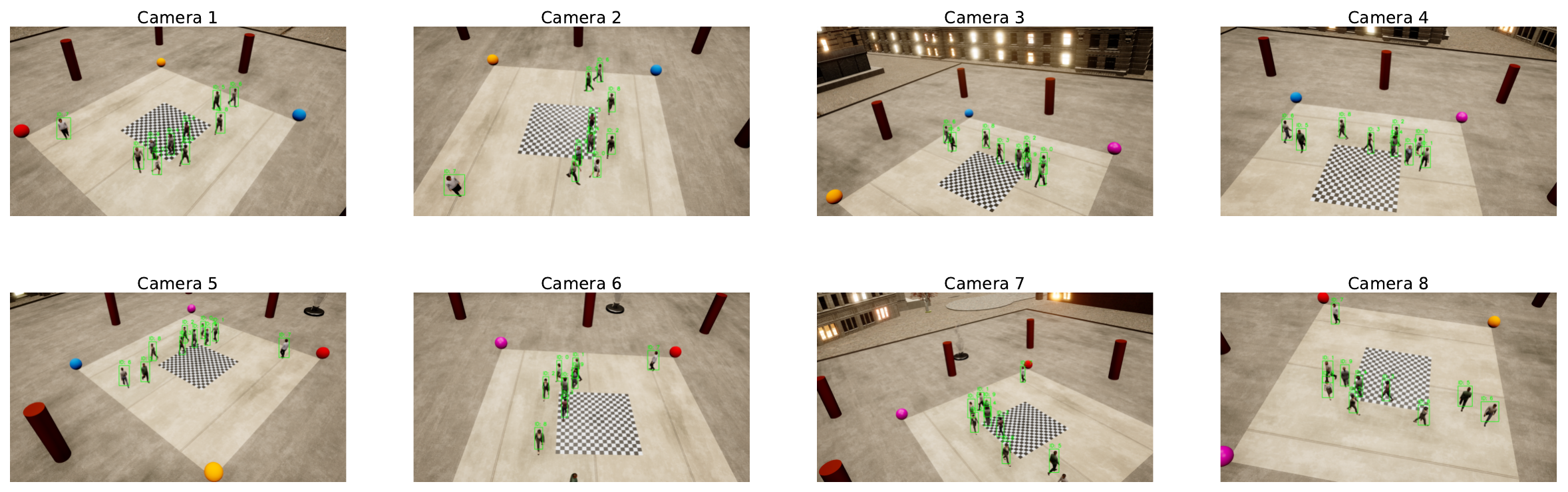}
	\caption{Sample synchronized snapshots from all eight drones at a single timestep in the simple MATRIX dataset. This scenario features 10 pedestrians in an open 15$\times$15 m environment with minimal occlusions and calibration checkerboards (checkered patterns) visible for dynamic camera calibration, providing a baseline for evaluating multi-drone tracking under favorable conditions. }
	\label{Fig_Combined_Images}
\end{figure*}

\begin{figure*}[ht]
    \normalsize
	\centering
   	\includegraphics[width=\textwidth]{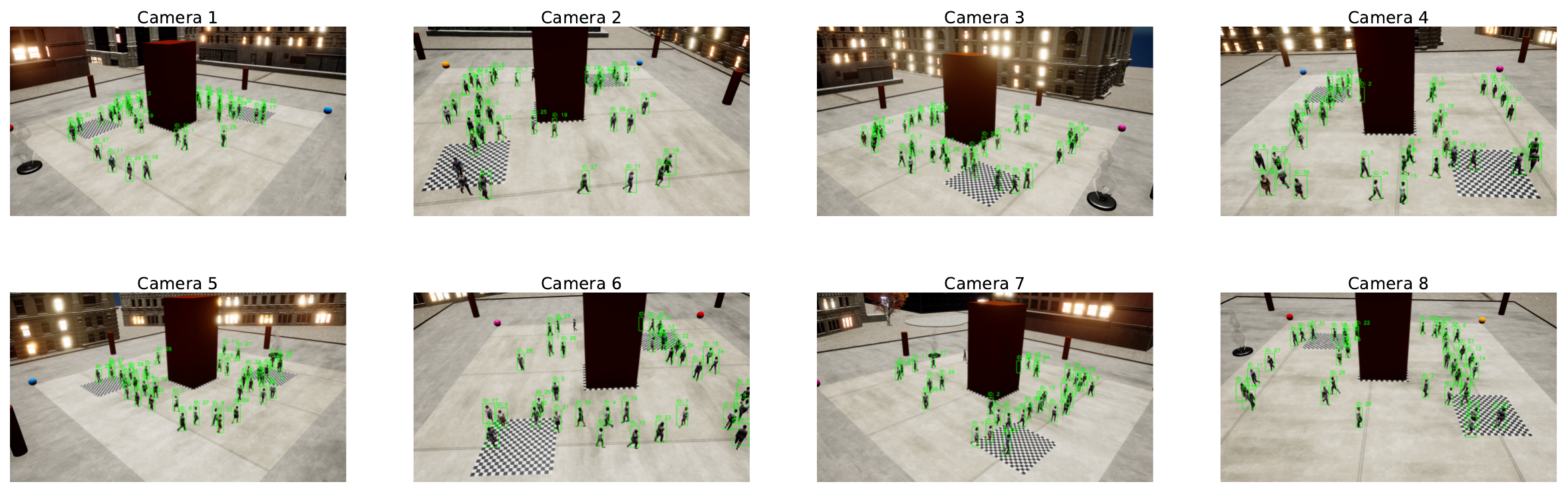}
	\caption{Sample synchronized snapshots from all eight drones at a single timestep in the complex MATRIX dataset. This scenario features 40 pedestrians in a 30$\times$30 m environment with a large central architectural obstruction (dark column) creating significant occlusions and dense crowding conditions that challenge tracking robustness. Note the varying perspectives and overlapping coverage areas across drone views.}
	\label{Fig_Combined_Images_Large}
\end{figure*}

\begin{table*}[t]
\caption{Drone Datasets}
\centering
\begin{tabular}{l c c c c}
\hline\hline
Name & Type & Resolution\footnotemark[1] & \#Frames\footnotemark[2] & Key Features \\
\hline
UAV123~\cite{Mueller2017} & Drone SOT & 1280$\times$720 & 110k & Diverse scenes \\
VisDrone-SOT~\cite{zhu2021detection} & Drone SOT & 1920$\times$1080 & 139.3k & Single object tracking \\
VisDrone-MOT~\cite{zhu2021detection} & Drone MOT & 1920$\times$1080 & 40k & Multi-object tracking \\
DroneVehicle~\cite{9759286} & Multi-Modal MOT & 1920$\times$1080 & 57k & RGB + Infrared \\
MDOT~\cite{9298794} & Multi-Drone SOT & 1920$\times$1080 & 260k & 2-3 drones, coordinated tracking \\
MDMT~\cite{10008047} & Multi-Drone MOT & 1920$\times$1080 & 40k & 2 drones, occlusion labels \\
MUMO~\cite{9913874} & Multi-Drone MOT & 1920$\times$1080 & 185k & 4 cameras, toy car tracking \\
\hline
\end{tabular}
\begin{tablenotes}
\centering
\item\footnotemark[1] Note, the resolution indicates the maximum resolution of videos/images included in the benchmarks and datasets
\item\footnotemark[2] k=1,000
\end{tablenotes}
\label{Drone_DS_Table}
\end{table*}

\begin{table*}[t]
\caption{Multi-View Datasets}
\centering
\begin{tabular}{l c c c c c}
\hline\hline
Name & \#Cameras & Resolution\footnotemark[1] & \#Frames & Target Type & Key Features \\
\hline
WildTrack~\cite{8578626} & 7 & 1920$\times$1080 & 400$\times$7 & Pedestrian & Real-world \\
MultiViewX~\cite{hou2020multiview} & 6 & 1920$\times$1080 & 400$\times$6 & Pedestrian & Game engine \\
GMVD~\cite{Vora_2023_WACV} & 3-8 & 1920$\times$1080 & 5995 & Pedestrian & Multiple scenes, weather variations \\
Synthehicle~\cite{10031087} & 340 & 1920$\times$1080 & 17 hours & Vehicle & 2D and 3D tracking \\
\textbf{MATRIX~(Ours)} & 8 & 1920$\times$1080 & 1000$\times$8 & Pedestrian & Moving drone views \\
\hline
\end{tabular}
\begin{tablenotes}
\centering
    \item\footnotemark[1] Note, the resolution indicates the maximum resolution of videos/images included in the benchmarks and datasets
\end{tablenotes}
\label{MV_DS_Table}
\end{table*}

\section{MATRIX Dataset}~\label{Section_III}

The MATRIX (Multi-Aerial TRacking In compleX environments) dataset represents a significant advancement in multi-view pedestrian tracking, offering synchronized footage from multiple aerial perspectives in a complex urban environment. Fig.~\ref{Fig_Combined_Images} and Fig.~\ref{Fig_Combined_Images_Large} show concurrent views from eight drones from the simple and complex dataset respectively, providing comprehensive coverage of the surveillance area. This section details the dataset's creation process, characteristics, and annotation methodology.

\subsection{Dataset Generation Pipeline}
The MATRIX dataset is generated through an advanced pipeline that integrates Unreal Engine 5~\cite{unrealengine5} with AirSim~\cite{airsim} for precise drone control and data collection. The generation process consists of several synchronized components that work together to create a comprehensive multi-view surveillance dataset:

\subsubsection{Simulation Environment Setup}

\begin{figure*}[!t]
{\centering
\includegraphics[width=\linewidth]{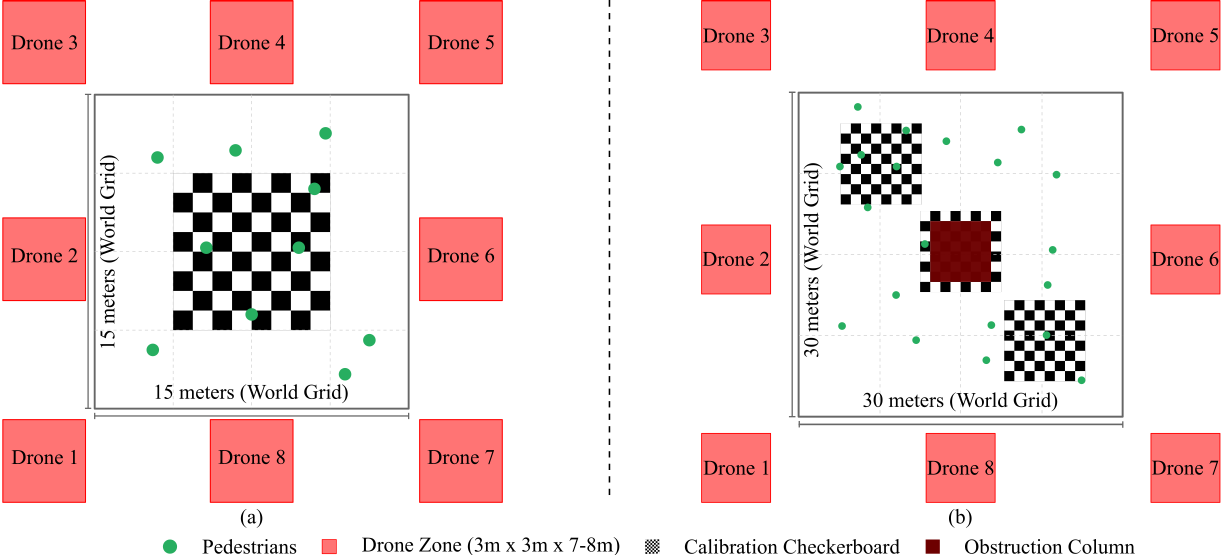}
\caption{An illustration of the Unreal Engine environment, where (a) is the smaller environment and (b) is the more complex test environment with occlusion. }
\label{Fig_Scene}}
\footnotesize
\end{figure*}

The two test environments, illustrated in Fig.~\ref{Fig_Scene}, are constructed using Unreal Engine 5, featuring:
\begin{itemize}
\item An urban landscape with realistic buildings
\item Dynamic pedestrian agents with realistic crowd behavior
\item Eight drones equipped with high-resolution cameras (1920$\times$1080 pixels) with 70$\degree$~field of view
\item Ideal weather and lighting conditions
\item Programmable flight paths within designated airspace boundaries of $\pm$3~m in $x$ and $y$ directions, and 7--8~m in $z$ direction
\end{itemize}

\subsubsection{Drone Control System}
Each drone's camera movement is managed through a sophisticated control system. PID controllers are utilized for camera focusing on the center of the scene~\cite{Ziegler1942OptimumSF}
\begin{equation}
    u(t) = K_\mathrm{p} e(t) + K_\mathrm{i} \int_0^t e(\tau)d\tau + K_\mathrm{d} \frac{d}{dt}e(t) ,
\end{equation}
where the gains $K_\mathrm{p}=0.0002$, $K_\mathrm{i}=0.00002$, and $K_\mathrm{d}=0.0001$ were empirically tuned to achieve smooth camera motion while maintaining responsive tracking of the scene center.
    
To control the movement of the drones, velocity-based position control~\cite{1087247} is utilized
\begin{equation}
    v = \frac{p_\text{target} - p_\text{current}}{||p_\text{target} - p_\text{current}||} \cdot v_\text{max} ,
\end{equation}
where $v_\text{max}$ is set to 1.5 m/s to ensure stable flight dynamics and prevent abrupt movements that could affect image quality.

\subsubsection{Real-Time Camera Calibration}
The data collection occurs at regular intervals (0.5 seconds) and includes drone camera calibration, for which we utilize a checkerboard pattern strategically placed in multiple areas of the scene. This approach allows us to obtain precise 3D coordinates of each checkerboard intersection, which are then matched with their corresponding 2D locations in the images captured by each drone camera. The calibration process yields both intrinsic and extrinsic camera parameters. Given a set of 3D world points $X_i$ and their corresponding 2D image points $x_i$, we solve the following equation~\cite{Hartley2004}
\begin{equation}
    x_i = K[R|\mathbf{t}]X_i ,
\end{equation}
where $K$ is the intrinsic matrix, and $[R|\mathbf{t}]$ represents the extrinsic parameters (rotation and translation).

\subsection{Ground Truth Generation and Annotation}
The generation of accurate ground truth data for multi-view pedestrian tracking presents unique challenges, particularly in scenarios with moving cameras and frequent occlusions. Our pipeline addresses these challenges through a comprehensive approach combining POMs with precise 3D position tracking and automated annotation generation.

\subsubsection{Probabilistic Occupancy Maps}
POMs serve as a crucial intermediate representation that bridges the gap between raw camera inputs and final annotations. These maps discretize the ground plane into a regular grid and estimate the probability of pedestrian presence at each location~\cite{4359319}
\begin{equation}
\text{POM}(x,y) = P(\text{occupied}|I_{1},...,I_{N}) ,
\end{equation}
where $I_{1},...,I_{N}$ represent views from N cameras. 

\subsubsection{Line-of-Sight Verification}
For each drone camera position $C = (x_c, y_c, z_c)$ and pedestrian position $P = (x_p, y_p, z_p)$, the line-of-sight verification involves calculating a ray vector $\vec{r}$, which physically represents the straight-line path that light travels from the pedestrian to the camera sensor, and checking for intersections with scene geometry along this path. The ray is defined as~\cite{ROTH1982109}
\begin{equation}
\vec{r}(t) = C + t(P - C), \quad t \in [0, 1] ,
\end{equation}
where $t$ is a dimensionless parameter that varies from 0 (at the camera position) to 1 (at the pedestrian position).

The LoS check determines if there are any intersections between $\vec{r}(t)$ and scene geometry (buildings, obstacles) for $t \in [0, 1]$. A binary visibility flag $v$ is assigned
\begin{equation}
v = \begin{cases}
1 & \text{if no intersections exist} \\
0 & \text{if any intersection is found} .
\end{cases}
\end{equation}

This calculation is performed for each drone-pedestrian pair at every timestep using Unreal Engine's ray-casting API, which efficiently computes geometry intersections. The resulting visibility matrix informs the annotation process by filtering detections to include only valid line-of-sight observations.

\subsubsection{Automated Annotation Pipeline}
The annotation system generates comprehensive metadata for each frame through multiple coordinated processes. For 2D annotations, the system automatically generates bounding boxes for all visible pedestrians, incorporating visibility flags based on precise line-of-sight calculations and providing sub-pixel accurate coordinates for maximum precision. The system also assigns unique person IDs to each detected individual and maintains these identities consistently across temporal sequences. Cross-view identity association ensures that individuals are correctly identified across multiple drone perspectives, even when they move between different camera fields of view or become temporarily occluded. 

The final annotation output encompasses several key elements that make the dataset particularly valuable for research and development. Each frame contains 2D bounding boxes with associated confidence scores, alongside precise 3D world coordinates for every detected person. The system maintains binary visibility flags for sophisticated occlusion handling. Temporal tracking information with consistent IDs allows for long-term movement analysis, while detailed camera calibration parameters for each frame enable accurate multi-view analysis. This comprehensive annotation approach ensures that researchers have access to all necessary information for developing and evaluating advanced multi-view tracking algorithms.

\section{Multi-View Detection and Tracking Framework}~\label{Section_IV}

\begin{figure*}[ht]
    \normalsize
	\centering
   	\includegraphics[width=\textwidth]{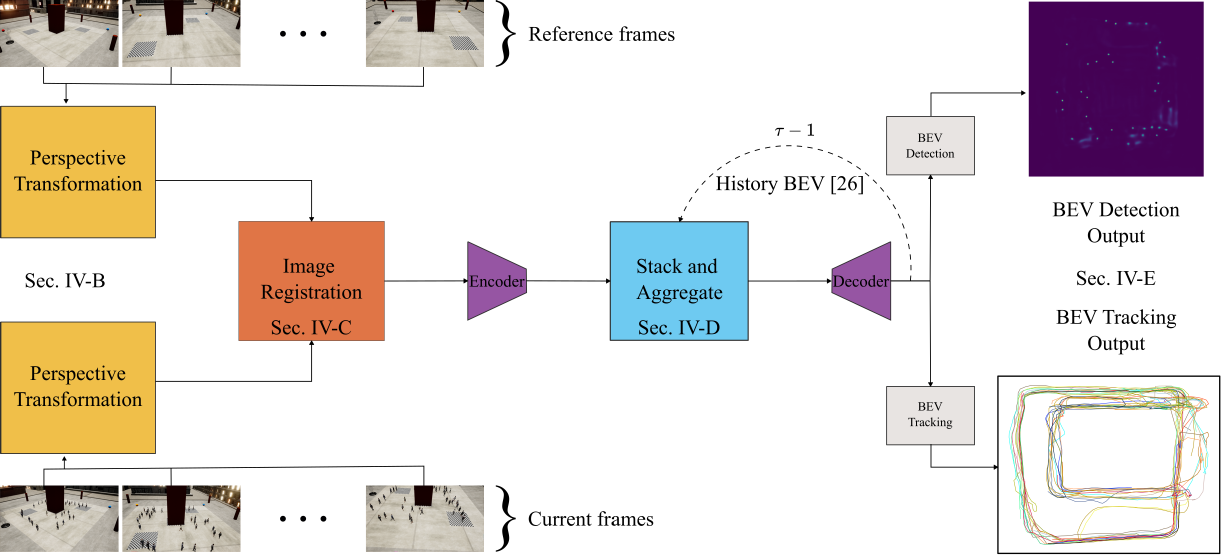}
	\caption{Illustration of the multi-view detection and tracking pipeline. The system processes synchronized frames from multiple drone cameras through two parallel streams: a reference stream that maintains a stable view representation, and a current stream that processes incoming frames. Each stream undergoes perspective transformation to BEV representation, followed by image registration that aligns the current view with the reference using feature matching and homography estimation. The registered multi-view features are then fused spatially and temporally before being decoded into detection heatmaps and tracking predictions. This dual-stream architecture enables the system to maintain spatial consistency despite continuous drone movement while preserving temporal coherence for robust tracking.}
	\label{Fig_Pipeline}
\end{figure*}

This section presents our comprehensive approach to multi-view pedestrian detection and tracking using dynamic drone cameras. Our methodology addresses the key challenges of dynamic viewpoint registration and occlusion handling through a novel pipeline that combines perspective transformation, feature-based image registration, and multi-view feature fusion.

\subsection{Dynamic Projection Matrix Updates}~\label{dynamic_bev}
To address the challenge of continuously changing drone positions, our pipeline updates the projection matrices $P'^{(s)}_t$ at each timestep. As each drone moves to a new position, the extrinsic parameters $[R|\mathbf{t}]$ change, requiring real-time recalibration. Our system leverages the strategically placed checkerboard patterns in the scene to continuously estimate new camera poses through perspective-n-point (PnP) solutions~\cite{Hartley2004}. The updated projection matrix for drone $s$ at time $t$ is computed as

\begin{equation}
P'^{(s)}_t = K^{(s)} [R_t^{(s)}|\mathbf{t}_t^{(s)}]_{3 \times 4} ,
\end{equation}

where $R_t^{(s)}$ and $\mathbf{t}_t^{(s)}$ represent the rotation and translation parameters for drone $s$ at time $t$. This ensures that the perspective transformation in Equation (10) accounts for the current drone position, enabling accurate BEV projection despite camera movement.

\subsection{Multi-View Detection and Tracking Pipeline}
Building upon the dynamic calibration framework, our pipeline consists of several interconnected components as illustrated in Fig.~\ref{Fig_Pipeline}, which are designed to process multi-view drone footage effectively. The system begins by receiving synchronized frames from multiple drone cameras, each providing a unique perspective of the surveillance area. These feeds undergo parallel processing through three main stages: perspective transformation to achieve a unified BEV representation, image registration to align current frames with reference views, and feature fusion to combine information from multiple viewpoints.

The pipeline employs a two-stream architecture where one stream processes the current frame while the other maintains reference information. This dual-stream approach enables robust handling of viewpoint changes while maintaining temporal consistency. The output of the pipeline provides both 2D detections in the image plane and localizations in worldgrid coordinates in the BEV space, along with consistent tracking identities across frames.

\begin{figure}[!t]
{\centering
\includegraphics[width=\linewidth]{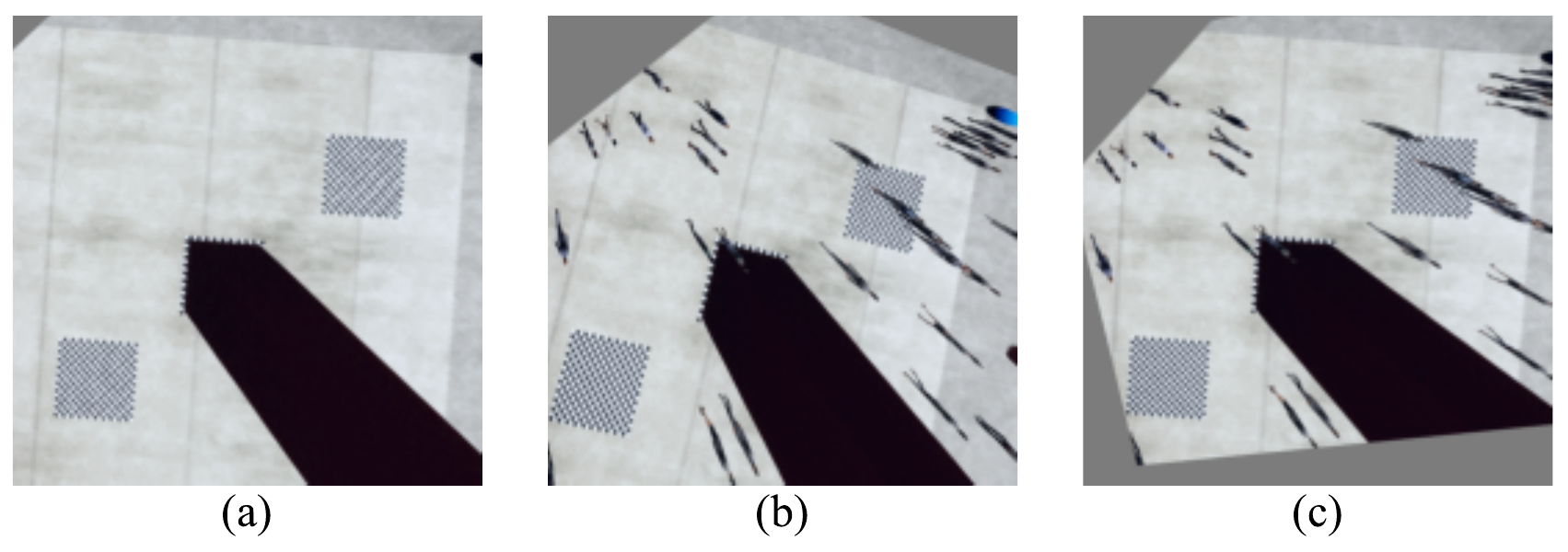}
\caption{Visualization of the image registration process, where (a) is the reference image, (b) is the original image from the batch, and (c) is the output after image alignment. }
\label{Fig_Registration}}
\footnotesize
\end{figure}

\subsection{Perspective Transformation}
The perspective transformation module converts each drone's view into a unified BEV representation, enabling consistent spatial reasoning across multiple viewpoints. Unlike static camera systems, our approach must handle continuously changing projection matrices as drones move through the surveillance space.

The perspective transformation uses the updated projection matrices from Sec.~\ref{dynamic_bev} to map features from different camera views onto a common ground plane reference. For each drone $s$ at timestep $t$, the transformation from 3D world coordinates $(x, y, z)$ to 2D image pixel coordinates $(u, v)$ follows the standard pinhole camera model~\cite{Hartley2004}

\begin{equation}
s\begin{bmatrix}
u \\
v \\
1
\end{bmatrix} = P_t^{(s)}\begin{bmatrix}
x \\
y \\
z \\
1
\end{bmatrix} = K^{(s)} [R_t^{(s)}|\mathbf{t}_t^{(s)}]\begin{bmatrix}
x \\
y \\
z \\
1
\end{bmatrix},
\end{equation}

where $s$ is a scaling factor, $P_t^{(s)}$ is the $3 \times 4$ camera projection matrix for drone $s$ at time $t$, $K^{(s)}$ is the intrinsic camera matrix, and $[R_t^{(s)}|\mathbf{t}_t^{(s)}]$ represents the time-varying extrinsic parameters.

For our BEV projection, we assume all points of interest lie on the ground plane $(z = 0)$, which simplifies the projection to a planar homography. This assumption is valid for pedestrian tracking as we focus on foot positions on the ground plane. The simplified projection becomes

\begin{equation}
s\begin{bmatrix}
u \\
v \\
1
\end{bmatrix} = H_t^{(s)}\begin{bmatrix}
x \\
y \\
1
\end{bmatrix},
\end{equation}

where $H_t^{(s)}$ is the $3 \times 3$ homography matrix for drone $s$ at time $t$, derived by removing the third column from $P_t^{(s)}$ (corresponding to the $z$-coordinate). This homography matrix is updated at each timestep as the drone moves, ensuring accurate ground plane projection despite camera motion.

The dynamic nature of our approach requires continuous updates to these homography matrices, but this enables robust BEV projection across the entire surveillance sequence. We apply this time-varying transformation to project features from all $S$ drones onto a predefined ground plane grid of size $[H_g, W_g]$, resulting in BEV features of size $S \times C_f \times H_g \times W_g$.

The key advantage of this approach is that while individual homography matrices $H_t^{(s)}$ change with drone movement, the unified BEV representation maintains consistent spatial relationships, enabling effective multi-view feature fusion despite the dynamic camera platform.

\subsection{Image Registration}
The image registration module, as shown in Fig.~\ref{Fig_Registration}, aligns current frames with reference views to maintain spatial consistency despite drone movement. Our approach utilizes the GFTT-AffNet-HardNet detector for robust feature extraction and matching. The registration process consists of three main steps:

\subsubsection{Feature Detection and Description}
For each image pair (current and reference), we extract keypoints $K$ and their descriptors $D$ using

\begin{equation}
    K = \{k_{1}, ..., k_{n}\} \in \mathbb{R}^{2 \times n} ,
\end{equation}

\begin{equation}
    D = \{d_{1}, ..., d_{n}\} \in \mathbb{R}^{d \times n} ,
\end{equation}

where $n$ is the number of detected keypoints and $d$ is the descriptor dimension.

\begin{figure*}[!t]
{\centering
\includegraphics[width=\linewidth]{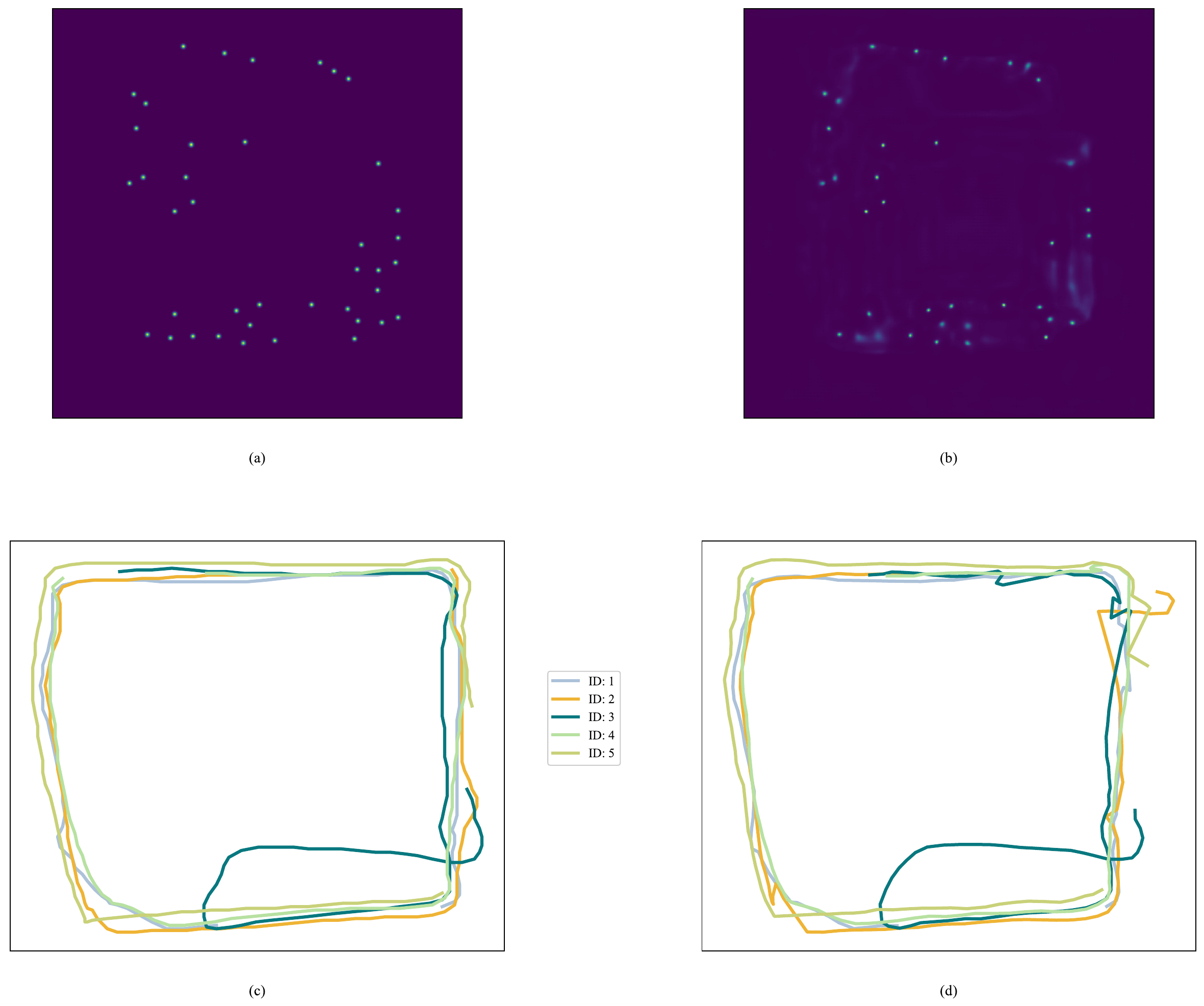}
\caption{Comprehensive visualization of detection and tracking performance. (a) Ground truth detection heatmap showing actual pedestrian locations from BEV. (b) Predicted detection heatmap demonstrating the model's ability to identify pedestrian positions. (c) Ground truth trajectories of tracked pedestrians, with each color representing a unique pedestrian ID. (d) Predicted tracking trajectories showing the model's tracking results. For clarity, only the five are displayed, selected from the total 40 tracked pedestrians. }
\label{Fig_det_and_track_visualization}}
\footnotesize
\end{figure*}

\subsubsection{Feature Matching}
Feature matching establishes correspondences between keypoints detected in the current frame and those in the reference frame. We employ a symmetric nearest neighbor matching strategy with Lowe's ratio test~\cite{lowe2004distinctive} to ensure robust matches while filtering out ambiguous correspondences.

For each descriptor $d_{i}$ in the current frame, we find its two nearest neighbors in the reference frame based on Euclidean distance. A match is considered valid if the ratio of distances to the first and second nearest neighbors falls below a threshold $\tau$

\begin{equation}
    M = \{(k_{i}, k_{j}) | \frac{d(d_{i}, d_{j})}{\min_{k \neq j} d(d_{i}, d_{k})} < \tau\} .
\end{equation}

where $d(d_{i}, d_{j})$ represents the Euclidean distance between descriptors $d_{i}$ and $d_{j}$, and $\tau$ is typically set to 0.8. This ratio test effectively rejects matches where multiple keypoints have similar descriptors, reducing false correspondences. 

To further improve robustness, we apply symmetric matching, where a correspondence $(k_{i}, k_{j})$ is only retained if $k_{j}$ is the best match for $k_{i}$ in the reference frame and $k_{i}$ is the best match for $k_{j}$ in the current frame. This bidirectional verification significantly reduces outliers in the final match set $M$.

\subsubsection{Homography Estimation}
The final transformation matrix $H$ is computed using RANSAC (Random Sample Consensus)~\cite{ransac} 
\begin{equation}
    H = \argmin_H \sum_{(p_i, p_j) \in M} \|p_j - Hp_i\|^2 ,
\end{equation}

The registration quality is assessed using a confidence score
\begin{equation}
    c = \frac{|\text{inliers}|}{|M|} ,
\end{equation}
where $|\text{inliers}|$ represents the number of matches that conform to the estimated homography, and $|M|$ is the total number of initial matches.

The stochastic nature of RANSAC introduces variability in the homography estimation across different runs, as the algorithm randomly samples match subsets to find the best transformation. This inherent randomness propagates through the pipeline, resulting in slight variations in detection and tracking performance between runs. To account for this variability, all results for our proposed method are reported with standard deviations computed
over multiple runs with different random seeds.

\subsection{Stack and Aggregate}
Our network architecture combines spatial and temporal information through a hierarchical fusion strategy. The approach consists of three conceptual stages that transform raw multi-view imagery into coherent detection and tracking outputs.

First, a convolutional encoder extracts visual features from each drone's perspective independently. These features capture appearance information while preserving spatial relationships within each view. Second, we aggregate features across views by stacking them in BEV space and applying a compression operation that reduces redundancy while preserving complementary information from different viewpoints. This multi-view fusion creates a unified representation that leverages the geometric consistency of the BEV projection.

Third, we incorporate temporal consistency by fusing the current BEV features with historical information from previous frames. This temporal aggregation enables the system to maintain track continuity and resolve ambiguities through motion cues. The fused spatio-temporal features are then decoded through specialized prediction heads that estimate object centers, sizes, orientations, and tracking embeddings. This architecture enables end-to-end learning of both detection and tracking within a single framework, where spatial fusion handles occlusions through multiple viewpoints and temporal fusion maintains identity consistency across frames.

\subsection{Output}

The network produces two primary types of outputs visualized in Fig.~\ref{Fig_det_and_track_visualization}:
\begin{itemize}
    \item BEV Detection Output: The detection heatmaps show the network's ability to localize pedestrians in BEV. Comparing (a) and (b) in Fig.~\ref{Fig_det_and_track_visualization}, we can observe the close correspondence between ground truth detections and predicted pedestrian locations. The detection output is generated through the center head of the decoder, which produces confidence scores for each grid cell in the BEV plane.
    \item BEV Tracking Output: The tracking visualization in Fig.~\ref{Fig_det_and_track_visualization} demonstrates the system's ability to maintain consistent object identities over time. The comparison between ground truth trajectories (c) and predicted tracks (d) shows how the network successfully leverages temporal information to maintain consistent object identities. The tracking output is generated by combining detection results with temporal offset predictions, enabling the system to associate detections across frames and form coherent trajectories.
\end{itemize}

\section{Experimental Results}~\label{Section_V}

This section presents a comprehensive evaluation of our proposed multi-view detection and tracking methodology. We conduct experiments on both variants of our MATRIX dataset to assess the effectiveness of our approach under different environmental conditions and compare against state-of-the-art methods.

\subsection{Evaluation Metrics and Datasets}

We employ established metrics from both detection and tracking literature to provide a thorough assessment of system performance. 

\textbf{Detection Metrics:}
\begin{itemize}
    \item {Multiple Object Detection Accuracy \textbf{(MODA)}}: Provides a comprehensive accuracy measure by considering both false positives and false negatives: $MODA = 1 - \frac{FP + FN}{GT}$, where $FP$, $FN$, and $GT$ represent false positives, false negatives, and ground truth objects, respectively.
    \item {Multiple Object Detection Precision \textbf{(MODP)}}: Evaluates the spatial precision of detections by measuring the overlap between predicted and ground truth bounding boxes.
\end{itemize}

\textbf{Tracking Metrics:}
\begin{itemize}
    \item {Multi-Object Tracking Accuracy \textbf{(MOTA)}}: Extends detection accuracy to incorporate identity switches: $MOTA = 1 - \frac{FP + FN + IDSW}{GT}$, where $IDSW$ represents identity switches.
    \item {Multi-Object Tracking Precision \textbf{(MOTP)}}: Measures the average distance between predicted and ground truth positions across all correctly matched detections.
    \item {ID F1 Score \textbf{(IDF1)}}: Evaluates identity preservation by measuring the ratio of correctly identified detections to the total number of detections and ground truth objects.
    \item {Mostly Tracked \textbf{(MT)}}: Quantifies the percentage of ground truth trajectories successfully tracked for at least 80\% of their duration.
\end{itemize}

\textbf{Dataset Variants:}
\begin{itemize}
    \item {Simple variant} (Fig.~\ref{Fig_Scene}(a)): Features 10 pedestrians in an unobstructed 15$\times$15 m environment, providing a baseline scenario for algorithm comparison.
    \item {Complex variant} (Fig.~\ref{Fig_Scene}(b)): Presents a more challenging 30$\times$30 m environment with 40 pedestrians and significant architectural occlusions, designed to test algorithm robustness under realistic surveillance conditions.
\end{itemize}

Both scenarios feature synchronized footage from eight drones with continuously changing positions, differentiating our dataset from existing static multi-view benchmarks. We compare our approach against EarlyBird~\cite{teepe2023earlybird} and TrackTacular~\cite{teepe2024lifting}, both recognized for strong performance on traditional multi-view datasets. All experiments use consistent training procedures with 100 epochs unless otherwise specified, ensuring fair comparison across methods. For our proposed method, we report mean performance metrics along with standard deviations (shown as error bars in figures) computed over 10 independent runs
with different random seeds. This accounts for the stochastic nature of the RANSAC algorithm used in our image registration module. Baseline methods (EarlyBird and TrackTacular) produce deterministic results and thus show no variation across runs.

\subsection{Baseline Performance}

\begin{table*}
\centering
\begin{threeparttable}
\caption{Stationary Dataset Performance}
\label{Table_SOTA}
\setlength{\tabcolsep}{4pt}
\begin{tabular}{@{}l*{10}{c}@{}}
\toprule
\multirow{2}{*}[-0.5ex]{Paper} & \multicolumn{4}{c}{Wildtrack} & \multicolumn{4}{c}{MultiviewX} \\
\cmidrule(lr){2-5}\cmidrule(lr){6-9}
& MODA~[\%] & MODP~[\%] & MOTA~[\%] & MOTP~[\%] &  MODA~[\%] & MODP~[\%] & MOTA~[\%] & MOTP~[\%] \\
\midrule
MVTT~\cite{10031058} & \textbf{94.1} & \underline{81.3} & - & - & \underline{95.0} & \textbf{92.8} & - & - \\
MVDet~\cite{hou2020multiview} & 88.2 & 75.7 & - & - & 83.9 & 79.6 & - & - \\
MVDeTr~\cite{zhu2021deformable} & \underline{91.5} & \textbf{82.1} & - & - & 93.7 & \underline{91.3} & - & - \\
EarlyBird\textsuperscript{1}~\cite{teepe2023earlybird} & 91.2 & 81.8 & {89.5} & \underline{86.6} & 94.2 & 90.1 & {88.4} & \underline{86.2} \\
TrackTacular~\cite{teepe2024lifting} & 91.8 & 79.8 & \textbf{89.6} & {81.7} & \textbf{95.9} & 89.2 & \textbf{91.4} & \textbf{86.7} \\
{REMP}~\cite{dakic2024} & 
{90.9} & {79.4} & {88.5} & {\textbf{86.8}}  & 
{89.9} & {90.5} & {81.0} & {85.8} \\
\midrule
\textbf{Proposed method}\textsuperscript{2} & 91.2 & 79.7 & \underline{89.7} & 81.2 & 93.0 & 90.3 & \underline{90.7} & \textbf{86.7} \\
\midrule
\end{tabular}
\begin{tablenotes}
 \item[1] EarlyBird performance measured with our Python implementation differs 1--2\% from the original MATLAB-based results reported in~\cite{teepe2023earlybird}. This table presents the original published results.
 \item[2] Our method performs comparably to TrackTacular on static camera datasets, where image registration provides no benefit since viewpoints remain fixed throughout sequences. A standard deviation for the proposed method is not documented since they are negligible on static camera datasets.
\end{tablenotes}
\end{threeparttable}
\end{table*}

To validate our approach on established benchmarks, we evaluated our method on the MultiviewX~\cite{hou2020multiview} dataset, built in a virtual environment with six calibrated cameras and 400 frames capturing multiple pedestrians with ground plane bounding box annotations, and on Wildtrack~\cite{8578626}, a real-world dataset with seven synchronized cameras and 400 frames with ground plane point annotations, as shown in Table~\ref{Table_SOTA}. Our proposed method achieves performance comparable to TrackTacular, with MODA scores of 91.2\% vs. 91.8\% on WILDTRACK and 93.0\% vs. 95.9\% on MultiviewX. This similarity in performance is expected, as both datasets feature stationary cameras with fixed positions throughout the sequences. In these scenarios, our image registration module essentially registers each image against itself, providing no meaningful transformation or alignment benefit since the camera viewpoints remain constant. The core network architecture remains very similar to TrackTacular, differing primarily in the addition of the image registration component, which adds computational complexity without modifying the input images or improving performance on static camera datasets. These results demonstrate that our method maintains competitive performance on traditional multi-view benchmarks while providing the additional capability to handle dynamic camera scenarios—a crucial advantage that becomes apparent in our MATRIX dataset results where continuous camera movement necessitates robust view alignment.

\begin{figure*}[!t]
{\centering
\includegraphics[width=\linewidth]{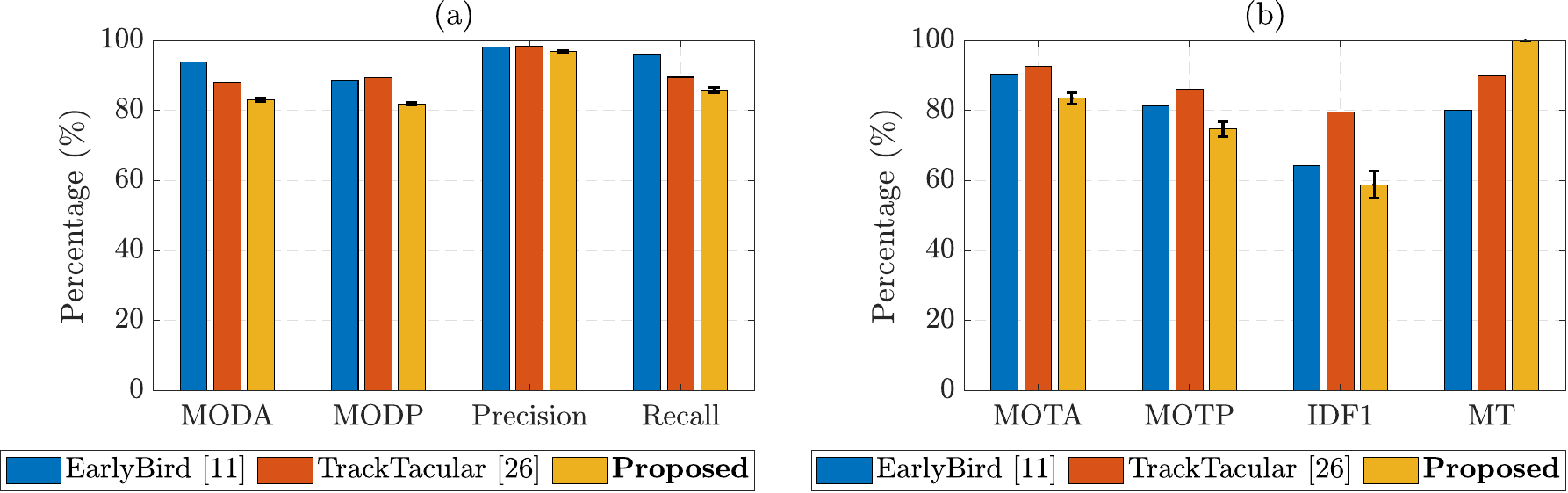}
		\caption{Detection performance (a) and tracking performance (b) on the simple MATRIX dataset variant (10 pedestrians, 15$\times$15 m). Error bars show standard deviation over 10 runs for the proposed method; baseline methods produce deterministic outputs.}
\label{Fig_simple_performance}}
\footnotesize
\end{figure*}

\begin{figure*}[!t]
{\centering
\includegraphics[width=\linewidth]{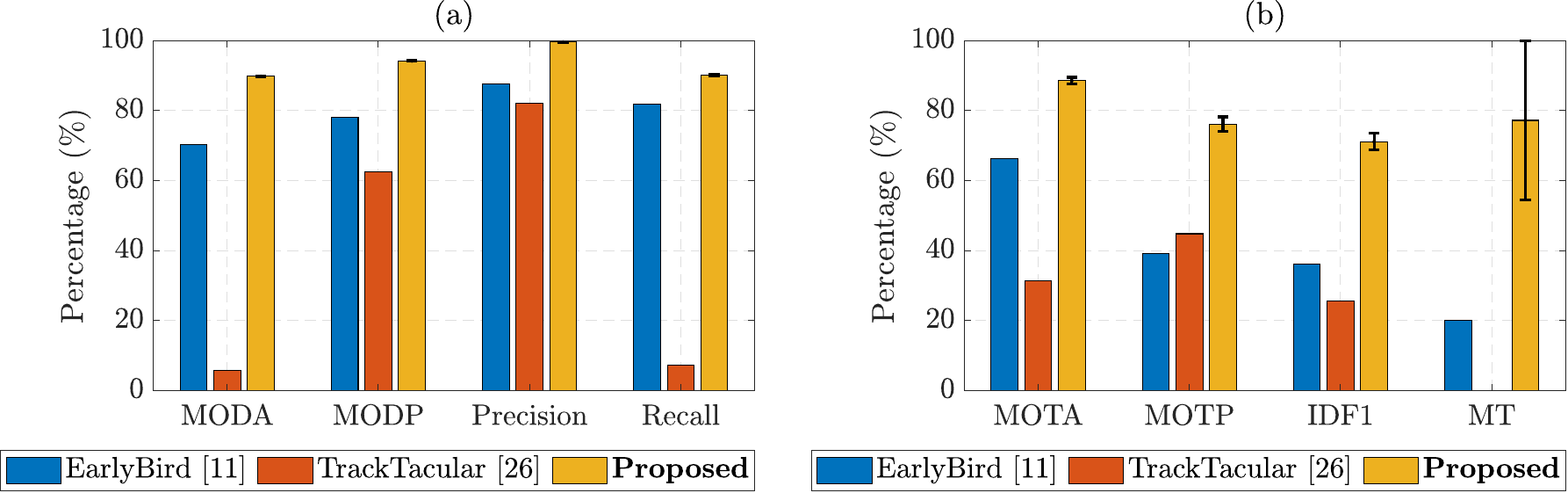}
		\caption{Detection performance (a) and tracking performance (b) on the complex MATRIX dataset variant (40 pedestrians, 30$\times$30 m, central obstruction). Error bars show standard deviation over 10 runs for the proposed method; baseline methods produce deterministic outputs.}
\label{Fig_complex_performance}}
\footnotesize
\end{figure*}

Fig.~\ref{Fig_simple_performance} and Fig.~\ref{Fig_complex_performance} present comprehensive performance comparisons between our proposed method and static camera approaches across both simple and complex MATRIX datasets, respectively. In the simple dataset scenario, which features only 10 pedestrians in an open environment, all three methods, including EarlyBird and TrackTacular and our proposed method demonstrate strong performance. EarlyBird achieves a notably high MODA of 94\% and recall of 95.9\% for detection, while TrackTacular excels in tracking with a MOTA of 92.7\% and an impressive IDF1 score of 79.64\%. This performance can be attributed to the relatively unchallenging nature of the simple dataset, where pedestrian density is low and the impact of drone movement on detection and tracking is minimal, reducing the necessity for sophisticated image registration techniques.

However, the complex dataset results in Fig.~\ref{Fig_complex_performance} reveal a significant performance disparity between the methods. TrackTacular's performance drastically deteriorates, with MODA dropping to 5.72\% and MOTA to 31.44\%, while achieving 0\% for mostly tracked trajectories. This substantial degradation can be attributed to the challenging nature of the complex environment, where increased pedestrian density combined with dynamic drone perspectives creates significant difficulties for traditional tracking approaches. In contrast, our proposed method maintains robust performance in the complex scenario, achieving $\sim$89\% MODA and $\sim$88.5\% MOTA. This marked improvement can be primarily attributed to our image registration component, which effectively aligns multiple viewpoints and maintains spatial consistency despite drone movement. The high precision ($\sim$97\%) and recall ($\sim$86\%) in detection further demonstrate the effectiveness of our approach in handling dense crowds and dynamic perspectives.

These results underscore a crucial finding: while existing static camera methods perform admirably in simple scenarios, they struggle to maintain performance in more complex, realistic conditions with dynamic viewpoints and dense crowds. Our proposed method, though showing marginally lower performance in simple scenarios, demonstrates significantly superior robustness and scalability when confronted with more challenging conditions, validating the effectiveness of our image registration-based approach for real-world applications.

\subsection{Impact of Pretraining on Modified Environments}

\begin{figure}[!t]
{\centering
\includegraphics[width=\linewidth]{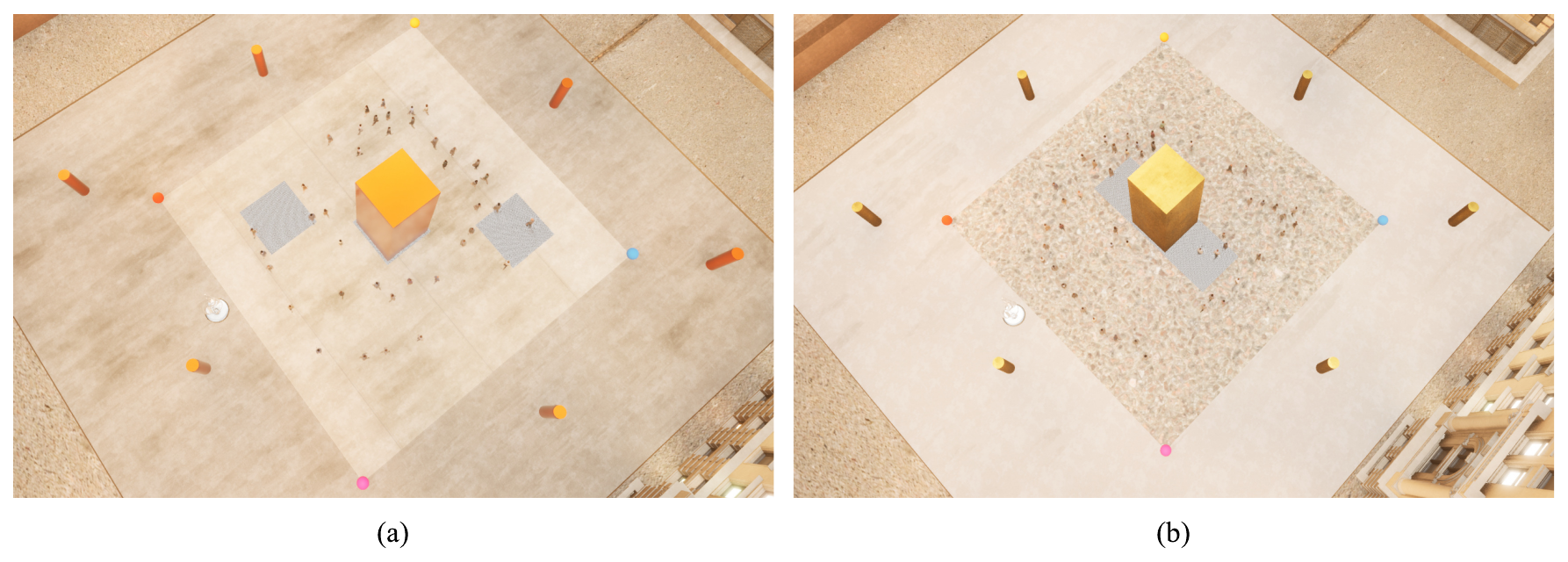}
\caption{Comparison of MATRIX environments for transfer learning evaluation. (a) Original complex environment with cement flooring and standard obstruction. (b) Modified environment with cobblestone flooring, altered obstruction appearance, and relocated calibration checkerboards, used to evaluate model generalization and pretraining benefits. Both scenarios maintain 30$\times$30 m dimensions but the modified version increases pedestrian density from 40 to 50 individuals. }
\label{Fig_Pretrain}}
\footnotesize
\end{figure}

\begin{figure}[!t]
{\centering
\includegraphics[width=\linewidth]{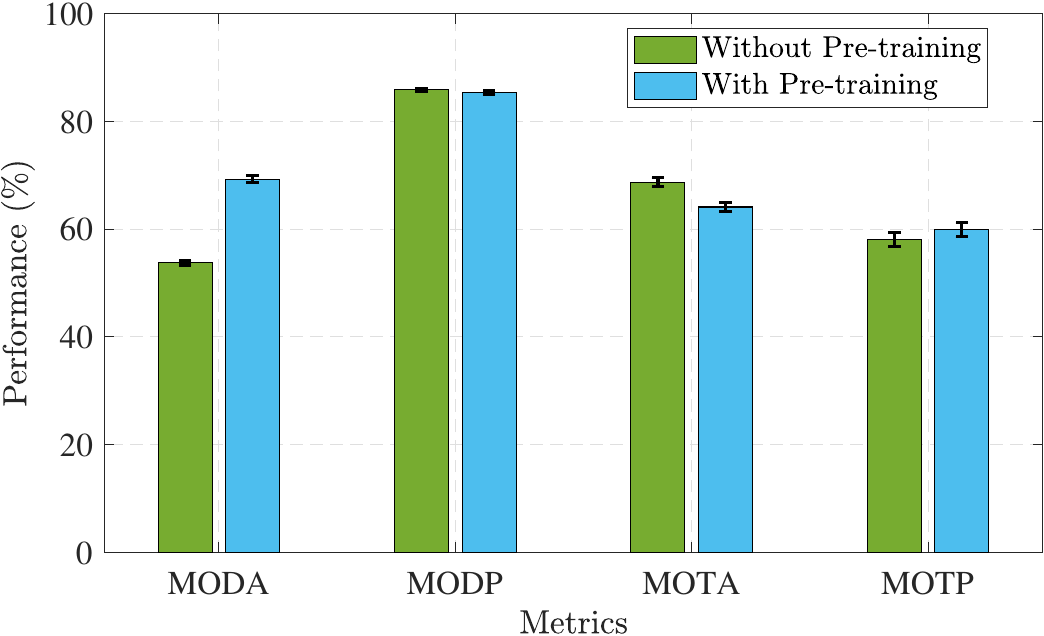}
\caption{Impact of transfer learning on modified MATRIX environment. Comparison of detection (MODA, MODP) and tracking (MOTA, MOTP) metrics between training from scratch (green) versus initializing with weights pretrained on the original complex dataset (blue). Both models trained on a modified dataset with only 50\% of the samples compared to the original. }
\label{Fig_pre_performance}}
\footnotesize
\end{figure}

To evaluate the generalization capabilities of our approach and investigate the benefits of transfer learning, we conducted experiments on a modified variant of the complex MATRIX dataset. This variant introduced environmental modifications, including altered surface materials (changing the floor from cement to cobblestone) and modifications to the obstruction material, changing the locations of the calibration checkerboard, while also increasing pedestrian density from 40 to 50 individuals. A vizualization of these environments is shown in Fig.~\ref{Fig_Pretrain}. We compared two training approaches: training from scratch using half of the modified dataset, and training with the same limited data but initializing with weights pretrained on the original complex MATRIX dataset.

The results, illustrated in Fig.~\ref{Fig_pre_performance}, demonstrate significant performance improvements in the MODA metric when utilizing pretrained weights. The pretrained model achieved notably higher performance, with MODA increasing from $\sim$54\% to $\sim$69\% (a 15\% point improvement) while maintaining consistently high MODP scores above 80\%. This substantial gain in detection accuracy demonstrates the model's ability to leverage learned feature representations from the original environment. Interestingly, the tracking metrics (MOTA and MOTP) remained relatively similar between the pretrained and from-scratch approaches, suggesting that while pretraining significantly enhances the model's ability to detect pedestrians in novel environments, the temporal association and identity maintenance capabilities require more environment-specific learning.

These results highlight several important findings about transfer learning in multi-view drone surveillance. First, pretraining on simulation data provides robust feature representations that generalize well to modified environments, even with limited training data. The 15\% improvement in MODA demonstrates that the model successfully transfers low- and mid-level visual features related to pedestrian appearance and spatial reasoning. Second, the ability to achieve strong performance with only half the training data addresses a critical practical concern for real-world deployments, where collecting and annotating large-scale datasets in every new environment is often prohibitively expensive. Third, the similar tracking performance between approaches suggests that while detection benefits substantially from transfer learning, tracking requires additional domain-specific adaptation, potentially due to differences in pedestrian movement patterns and occlusion characteristics between environments. This insight points to future research directions in developing more transferable temporal modeling components.

\subsection{Camera Dropout}

\begin{figure}[!t]
{\centering
\includegraphics[width=\linewidth]{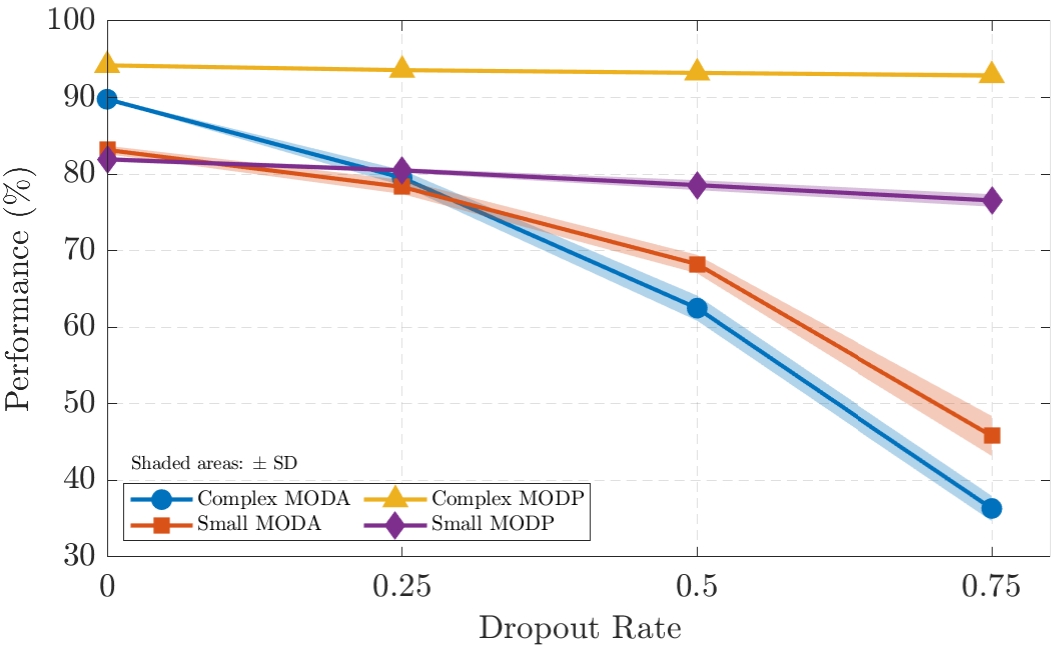}
\caption{Impact of camera dropout on detection performance. The graph shows MODA and MODP metrics as the dropout rate increases from 0 to 0.75 for both simple and complex MATRIX datasets. }
\label{Fig_dropout_detection}}
\footnotesize
\end{figure}

\begin{figure}[!t]
{\centering
\includegraphics[width=\linewidth]{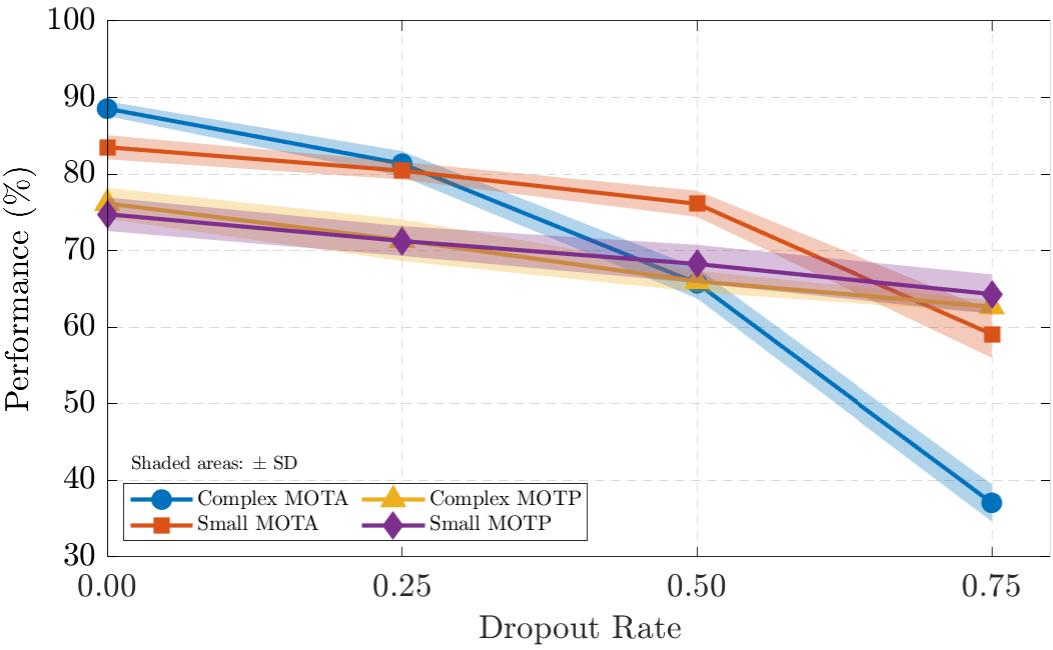}
\caption{Impact of camera dropout on tracking performance. The graph illustrates MOTA and MOTP metrics across varying dropout rates for both simple and complex MATRIX datasets.  }
\label{Fig_dropout_tracking}}
\footnotesize
\end{figure}

To evaluate the robustness of our approach to camera failures or limited deployment scenarios, we conducted systematic camera dropout experiments on both MATRIX dataset variants. This analysis provides insights into the minimum number of cameras required to maintain acceptable performance and how gracefully the system degrades as cameras become unavailable.

Fig.~\ref{Fig_dropout_detection} demonstrates the effect of camera dropout on detection performance, measured through MODA and MODP metrics. For both datasets, MODA maintains relatively high performance (>80\%) at 25\% dropout ($\sim$6 cameras), then degrades more noticeably at higher dropout rates, reaching approximately 45\% and 36\% on the simple and complex MATRIX dataset variants respectively, with 75\% dropout ($\sim$2 cameras). Comparing the complex dataset MODA performance to the simple variant shows that the completed dataset performance follows a similar pattern but with steeper degradation. MODP remains relatively stable across dropout rates for both datasets.

Fig.~\ref{Fig_dropout_tracking} illustrates the impact on tracking performance through MOTA and MOTP metrics. The complex dataset demonstrates solid tracking robustness, maintaining MOTA above 80\% up to 25\% dropout before declining to approximately 40\% at 75\% dropout. The simple dataset follows a similar trajectory but with slightly absolute values, where the MOTA declines to approximately 60\% at 75\% dropout. Both datasets maintain consistent MOTP performance across all dropout rates, with values remaining within a narrow band around 75-85\%, indicating that the spatial precision of tracked objects remains high regardless of the number of available cameras.

These results reveal several important insights about system robustness. First, the multi-view framework exhibits graceful degradation, with no catastrophic failure even at 75\% dropout rates. Second, the complex environment demonstrates that our approach can maintain reasonable performance ($\sim$60\% MOTA) even with only 2 cameras, though this represents a significant decrease from the MOTA achieved with all 8 cameras. Third, the maintenance of high precision metrics (MODP and MOTP) across dropout rates suggests that while fewer cameras reduce the system's ability to detect and track all targets, those that are successfully tracked maintain high positional accuracy. Finally, the relatively stable performance at a 25\% dropout rate indicates that the system can tolerate occasional camera failures without severe performance impact, an important consideration for real-world deployments where camera malfunctions may occur.

\section{Conclusion and Future Work}~\label{Section_VI}
This work addresses critical challenges in multi-view pedestrian detection and tracking for dynamic drone-based surveillance through the introduction of the MATRIX dataset and a novel multi-view framework. Our experimental evaluation demonstrates that while state-of-the-art static camera methods maintain over 90\% precision and accuracy in simple scenarios, their performance degrades dramatically in complex environments. In contrast, our proposed framework maintains robust performance with $\sim$90\% detection and tracking accuracy and successfully tracks $\sim$80\% of trajectories under challenging conditions featuring dense crowds, significant occlusion, and continuously moving camera platforms. The framework's dynamic camera calibration system and feature-based image registration enable real-time adaptation to changing viewpoints, while the multi-view feature fusion pipeline effectively handles occlusions through complementary perspectives.

Transfer learning experiments reveal strong generalization capabilities, with pretrained models achieving 15\% higher detection accuracy ($\sim$54\% to $\sim$69\%) on modified environments using only half the training data. Camera dropout analysis validates practical viability through graceful degradation: the system maintains >80\% MODA and MOTA with up to 25\% camera dropout, with precision metrics remaining consistently high across all dropout rates. This robustness addresses critical concerns for operational surveillance systems where camera failures are inevitable, demonstrating that multi-view drone systems can be designed with redundancy to tolerate occasional failures without catastrophic collapse. The MATRIX dataset, with its eight synchronized drone views and comprehensive annotations, establishes new benchmarks for evaluating tracking systems under realistic surveillance conditions.

The applications of this work span urban surveillance, crowd monitoring, event security, and search and rescue operations. Promising future directions include integrating additional sensor modalities (LiDAR, thermal imaging) for adverse condition robustness, real-world implementation studies on physical drone platforms, privacy-preserving techniques for ethical surveillance, efficient architectures for edge deployment, adaptive camera positioning strategies based on crowd density, and extensions to longer-term tracking with hundreds of targets. The MATRIX dataset provides a valuable benchmark for these advances, bridging the gap between controlled laboratory conditions and real-world deployment in dynamic multi-view tracking systems.

\bibliographystyle{ieeetr}
\bibliography{bibliography}
\end{document}